\documentclass{edm_article}
\usepackage{tabularx}
\usepackage{multirow}
\usepackage{rotating, booktabs}
\usepackage{multicol, blindtext}
\usepackage{hyperref}
\begin{document}
\title{Automatic Short Math Answer Grading \\ via In-context Meta-learning}
\numberofauthors{4} 
\author{
\alignauthor
Mengxue Zhang\\
        \affaddr{Massachusetts Amherst}\\
        \email{mengxuezhang@cs.umass.edu}
\alignauthor
Sami Baral\\
        \affaddr{Worcester Polytechnic Institute}\\
        \email{sbaral@wpi.edu}
\alignauthor Neil Heffernan\\
        \affaddr{Worcester Polytechnic Institute}\\
        \email{nth@wpi.edu}
\and  
\alignauthor Andrew Lan\\
        \affaddr{Massachusetts Amherst}\\
        \email{andrewlan@cs.umass.edu}
}

\maketitle

\begin{abstract}

Automatic short answer grading is an important research direction in the exploration of how to use artificial intelligence (AI)-based tools to improve education. Current state-of-the-art approaches use neural language models to create vectorized representations of students responses, followed by classifiers to predict the score. However, these approaches have several key limitations, including i) they use pre-trained language models that are not well-adapted to educational subject domains and/or student-generated text and ii) they almost always train one model per question, ignoring the linkage across question and result in a significant model storage problem due to the size of advanced language models. In this paper, we study the problem of automatic short answer grading for students' responses to math questions and propose a novel framework for this task. First, we use MathBERT, a variant of the popular language model BERT adapted to mathematical content, as our base model and fine-tune it on the downstream task of student response grading. Second, we use an in-context learning approach that provides scoring examples as input to the language model to provide additional context information and promote generalization to previously unseen questions. We evaluate our framework on a real-world dataset of student responses to open-ended math questions and show that our framework (often significantly) outperform existing approaches, especially for new questions that are not seen during training. 

\end{abstract}


\section{Introduction}

\sloppy
Automated scoring (AS) refers to the problem of automatically scoring student (textual) responses to open-ended questions with multiple correct answers, often utilizing various machine learning algorithms. AS approaches can potentially scale up human grading effort: by training on a small number of example scores provided by human experts, they can automatically score a large number of responses. With the advancement in online learning platforms in recent years, there has been a growing body of research around the development of AS methods. AS has been studied in many different contexts, including automated essay scoring (AES)~\cite{asap,aes} and automatic short answer grading (ASAG)~\cite{irtasag,chiasag}, which has been studied in various different subject domains~\cite{sami,pardos,erickson,ibmin, fowler2021autograding, biology1, transfer1, review1}. The majority of AS approaches follow two steps: First, obtaining a \emph{representation} of student responses, often using methods in natural language processing, and second, applying a \emph{classifier} on top of this representation to predict the score~\cite{attali2006automated,danielle}. Over the years, AS approaches have gradually shifted from classic text representations such as bag-of-words or human-crafted features~\cite{erater,iea,cohmetrix,aes,relevance,semantic} that are human-interpretable to more abstract representations based on pre-trained neural language models~\cite{cambium,mayfield2020should,taghipour2016neural,uto2020neural,yang2020enhancing}. 


In this paper, we focus on ASAG in one particular subject domain: Mathematics. Math questions, or questions that involve mathematical reasoning, are ubiquitous in many science, technology, engineering, and mathematics (STEM) subject domains. Recently, several works~\cite{sami,erickson} have studied ASAG for the responses students provide to math-based open-ended questions that (are often concise) include their reasoning or thinking process about a particular concept. As noted in prior work, a key technical challenge in this domain is that student responses to math-based open-ended questions often are a combination of text (natural language) and mathematical language (symbols, expressions, and equations). However, most existing pre-trained language models such as BERT~\cite{devlin2018bert} and GPT~\cite{gpt3} are not specifically designed for mathematical language. Therefore, existing approaches for math ASAG that do not address the mathematical language present in student responses~\cite{sami,erickson} may not be able to accurately represent student reasoning processes in their responses. On the other hand, existing methods that focus entirely on mathematical language \cite{mlp,forte,mx} cannot process natural language contained in open-ended responses. 

Another significant limitation of existing AS approaches is that, in most cases, we need to train a separate AS model for each question. In contexts such as AES where questions (essay prompts) may not have high similarities, this approach can often be effective. However, in other contexts where reading comprehension or reasoning is involved, multiple questions may be linked to each other through the background information provided. In the context of math questions, many questions share similar skills or are different parts of a multi-step question. Therefore, training a separate AS model for each question would result in models that can only identify typical patterns in student responses to each individual question but cannot really understand how to differentiate good responses from bad ones. It is likely that these models would not be able to generalize well to previously unseen questions, as noted in~\cite{pardos}. 
More importantly, training a separate AS model for each question may create a significant problem for model storage and management. This problem is especially significant for state-of-the-art AS approaches that fine-tune pre-trained language models that have millions of parameters.

\subsection{Contributions}

In this paper, we develop an ASAG framework for students' open-ended responses to math questions. Building on a grand prize winning solution \cite{nigel} to the National Assessment of Educational Progress (NAEP) Automated Scoring Challenge\footnote{\url{https://github.com/NAEP-AS-Challenge/info}}, our framework is based on fine-tuning a pre-trained BERT language model on actual student responses, with several main innovations:
\begin{itemize}
    \item First, we use MathBERT \cite{mathbertcrap}, a version of the popular BERT language model adapted to mathematical content, as our base model. This model is capable of understanding math symbols and expressions to some extend and help us obtain a better representation of open-ended student responses.
    \item Second, we leverage in-context learning ideas in NLP research \cite{chen2021meta,min2021metaicl} and develop an ASAG approach using on multi-task and meta-learning tools (that are popular machine learning tools to promote model generalizability). Specifically, we fine-tune MathBERT with a carefully designed input format that uses \emph{example responses and scores} as additional input (together with question and response texts) to provide additional context of each question. This input format helps us train a shared AS model across all questions and out-performs the current state-of-arts approach \cite{sami}. 
    \item Third, we show that meta in-context learning leads to highly generalizabile AS models. Our intuition on why our approach is highly effective is that, by explicitly using example responses and scores as input, we reduce the AS task to a \emph{similar response finding} task, which is easier for the model to learn. 
\end{itemize}
We evaluate our ASAG framework on a real-world dataset which contains students' solution processes to open-ended math questions and grades provided by teachers. Through a series of quantitative experiments, we show that our framework (sometimes significantly) outperforms existing approaches in terms of score prediction performance. More importantly, we show that our framework significantly outperforms existing approaches \cite{sami, pardos} (by up to 50\% on some metrics under some settings) when applied to questions that are previously unseen during training, using only a few scored examples for these new questions. 
Perhaps surprisingly, we found that MathBERT does not provide additional benefit on top of the original BERT model while the in-context fine-tuning setup is key to the excellent generalization performance. We also summarize observations from qualitative evaluations of scoring errors, discuss the limitations of our framework, and outline several avenues for future work. Our implementation is publicly available.\footnote{\url{https://github.com/kikumaru818/meta_math_scoring}}

\section{Related Work}

In recent years, there have been many developments in ASAG methods across various domains. Most of the prior works have focused on non-mathematical domains \cite{basu2013powergrading,chen2013automated} where student responses are purely textual. However, more AS works have started to focus on more specific domains that contain non-textual symbols, e.g., Math, Physics, Chemistry, Biology and Computer Science \cite{cohmetrix,aes,semantic}. In these domains, a combination of natural language processing methods for the representation of responses and  machine learning methods for score classification has shown promising results \cite{sami,erickson,leacock2003c,srikant2014system,taghipour2016neural}. 

Here, we discuss two recent AS works in the mathematical domain, \cite{sami} and \cite{pardos}, that are the most relevant to our research. The authors of \cite{sami} proposed a scoring approach for short-answer math questions using sentence-BERT (SBERT)-based representation of student responses. Compared to this approach, our approach differs in many aspects and we highlight the following: First, we use MathBERT, a model pre-trained on mathematical content to represent student responses, while the approach in \cite{sami} ignores mathematical language in student responses. Second, we use an in-context meta-training approach to train one AS model for all questions while the approach in \cite{sami} trains one AS model for each question, which likely limits its generalizability to previously unseen questions. 

The authors of \cite{pardos} proposed a similar scoring approach for short-answer critical reasoning questions that combines various pre-trained representations, including SBERT, with classifiers for AS. Instead of using only student responses as input to the classifiers, they also use a series of question context information such as question text, rubric text, and question cluster identifier. As a result, they showed that their AS approach can generalize to previously unseen questions. 
Compared to this approach, our approach mostly differs in two aspects:
First, we fine-tune MathBERT on actual student responses while the approach in \cite{pardos} leaves the pre-trained representations fixed, which likely limits the accuracy of their student response representations. 
Second, we use scoring examples as input to MathBERT in addition to question text to further provide the AS model context of the question, which further enhances the generalizability of our model to previously unseen questions in a \emph{few-shot} learning setting.

\section{Methodology}
In this section, we detail both the ASAG setup for math questions and our in-context meta-learning framework. 

\subsection{Problem Statement}
We treat math ASAG as a classification problem where our goal is to train a scoring model that is capable of generalizing to new, previously unseen questions using a few examples. This setting is well studied in machine learning, commonly referred to as few-shot learning \cite{gpt3, chen2021meta, min2021metaicl}, where the goal is to train robust models that excel at multiple tasks.
Formally, we have a set of questions $T = \{Q_1, Q_2, Q_3, \ldots, Q_n\}$, where each question $Q_i \in T$ can be seen as a classification task. Each question $Q_i$ comes with numerous graded, training examples: $\{e^1_i, e^2_i, \ldots \}$. Each example consists of multiple fields of information: $e^i_j = \langle q_{text}, q_{id}, x, y \rangle$, where $q_{text}$ is the textual statement of the question, $q_{id}$ is an unique question id, $x$ is the text of student's response and $y$ is the grade from the teacher. We study on two problem settings in this work: i) generalization to new responses and ii) generalization to new questions. 


\subsubsection{Generalization to new responses}
This problem setting follows from that used in prior work \cite{sami}: we train a scoring method on scored responses for all questions and test it on held-out responses. We treat this problem setting as supervised learning classification and learn a scoring model $f: x \mapsto \hat{y}$ that predicts an estimated score $\hat{y}$ for a student response $x$ with true score $y$ by minimizing a loss function $L(y,\hat{y})$. For each question $Q_i$, we split the corresponding scored responses into two subsets, $Q^{Train}_i$ and $Q^{test}_i$, such that $Q^{Train}_i \cup Q^{test}_i = Q_i$ and $Q^{Train}_i \cap Q^{test}_i = \emptyset$. Instead of treating each question separately and train a model for each, we train one unified model on the union of training datasets for all questions, i.e., $\bigcup_{i=1}^{|T|} Q^{train}_{i}$. 
We detail the scoring model and our in-context learning setup in Section~\ref{in-context}. 

Let $\theta$ represent the model parameters, the optimization objective $\mathcal{L}_i$ for question $i$ is simply the cross entropy, i.e., the negative log-likelihood loss
\[\mathcal{L}_i(\theta) = \sum_{j: (x_j^i, y_j^i) \in Q_i^{train}} [-\log p_\theta (y_j^i | x_j^i, \ldots)]. \]
We minimize the total objective that spans all questions
\[\mathcal{L}(\theta) = \sum_{i = 1}^{|T|} \mathcal{L}_i(\theta) \]
to learn the model parameters $\theta$. 

\subsubsection{Generalization to new questions}
This problem setting can be formulated as a few-shot (or zero-shot) classification problem: we train a scoring model on scored responses for some questions and test its generalization capability to student responses to held-out questions. We first split the set of questions $T$ into $T_{train}$ and $T_{test}$ such that $T_{train} \cup T_{test} = T$ and $T_{train} \cap T_{test} = \emptyset$. We train the scoring model on all scored responses for the training questions $\bigcup_{i=1}^{|T_{train}|} Q_{i}$. Let $\gamma$ represent the model parameters for this problem setting, the optimization objectives for each question and across all training questions change to
\[\mathcal{L}_i(\gamma) = \sum_{j: (x_j^i, y_j^i) \in Q_i} [-\log p_\gamma (y_j^i | x_j^i, \ldots)] \]
and
\[\mathcal{L}(\gamma) = \sum_{i = 1}^{|T_{train}|} \mathcal{L}_i(\gamma), \]
respectively.

At test time, we applied the trained model to new questions $Q_i \in T_{test}$ to see how it can adapt using few (or zero) scored examples for these new questions. We study two cases: i) we do not update the original model with gradient updates, i.e., $\gamma$ remains unchanged, which we call the \textbf{Meta} setting, and ii) we update $\gamma$ by backpropagating gradients calculated on a few scored responses for new questions, which we call the \textbf{Meta-finetune} setting.

\begin{figure*}
    \includegraphics[width=\textwidth]{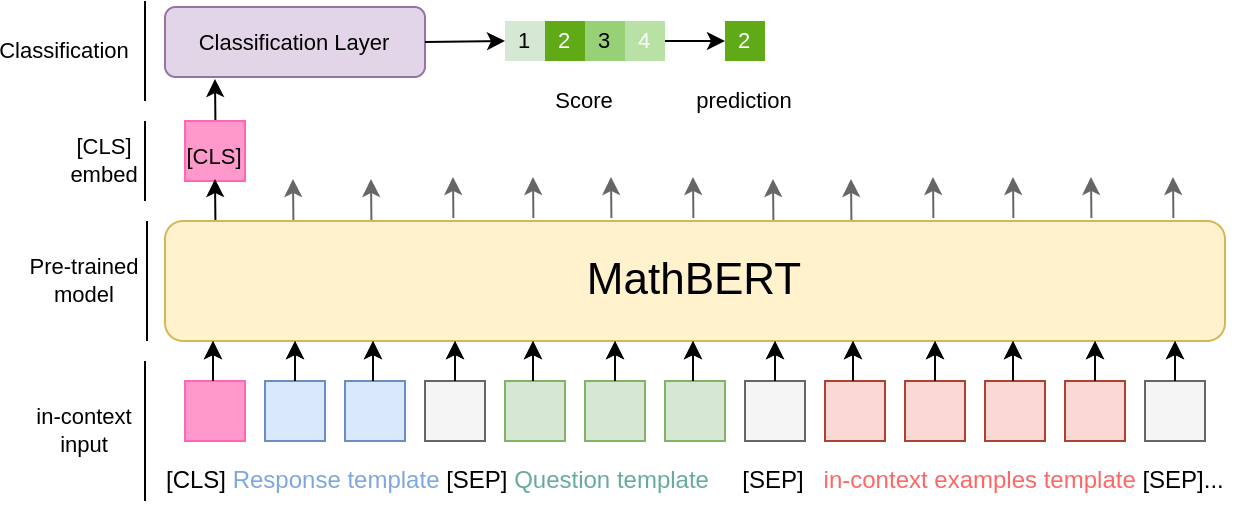}
    \caption{Overview of our in-context meta-learning-based ASAG method for math questions.}
    \label{fig:overview}
\end{figure*}

\begin{table*}[h]
\centering
\renewcommand{\arraystretch}{1.2}  
\caption{Templates for different components we use as input into our scoring method.}
\label{tab:template}
\resizebox{2\columnwidth}{!}{
\begin{tabular}{c|c|c} \toprule
Input Feature & Template & Sample Text \\ \hline 
Student Response & \textit{score this answer: $x^i_j$} & \textit{score this answer: expand the equation we get $2x+2 = 1$ then $x=-0.5$} \\ \hline
Question & \textit{question text: $q_{text_j}^i$ } & \textit{question text: Solve the equation $2(x+1)=1$} \\ \hline
Question ID & \textit{question id: $q_{id_j}^i$} & \textit{question id: 21314} \\ \hline
Scale &  \textit{scale: possible grade for question $i$ } & \textit{scale: poor, fair, good, excellent} \\ \hline  
Example & \textit{example: $x^i_{\neg j}$, score: $y^i_{\neg j}$} & \textit{example: move 2 to the right $x=1/2$, score: fair}  \\ \bottomrule
\end{tabular}}
\end{table*}

\subsection{BERT-based classification}
We now detail our scoring method based on fine-tuning a pre-trained language model. BERT \cite{devlin2018bert} is a pre-trained language model that produces contextualized representations of text and is also capable of encoding text. 
We use MathBERT \cite{mathbertcrap}, a variant of BERT pre-trained on a large mathematical corpus containing mathematical learning content ranging from pre-kindergarten (pre-k), high school, to college graduate levels. We use MathBERT as our base language model and fine-tune it on our data for downstream ASAG classification. 

Figure ~\ref{fig:overview} visualizes our method. The input to BERT is a sequence of tokens, starting with the [CLS] token, a special symbol added in front of every input during training process for BERT based model. Since [CLS] doesn't have meaning itself and BERT-based models learn contextualized representations of text, we can use the [CLS] embedding as a representation that encodes the entire input. 
We then feed the [CLS] embedding to a classification layer followed by softmax \cite{dlbook}, obtaining the predictive score class probabilities. A key difference between our work and prior works \cite{sami,pardos} that use BERT is that we also fine-tune the BERT model, i.e., update its parameters and adapt it to ASAG. Prior works only use BERT-based models to extract the representation of student responses; these methods are not likely going to be effective since they cannot adapt to student-generated content. During training, we backpropagate the gradient on the prediction objective to both i) the classification layer, which is learned from scratch, and ii) BERT, which is updated from its pre-trained parameter values.


\subsection{In-context Meta-learning}
\label{in-context}
Our key technical insight is that we need to use a well-crafted input format to provide context to the model and help it adapt to the scoring task for each question. 
Therefore, instead of only inputting the target student response we want to grade, we also include several other features as the input. These features are important to ground the model in the context of each question. 
For each possible feature, we also add additional textual instructions as input to the model about the semantic meaning of the feature. 

Table~\ref{tab:template} shows all possible features we include as model input and the corresponding template. Student response denotes the target responses to be scored. Thus, the corresponding textual instruction is ``score this answer.'' 
Since student responses are essential to the grading task, we place it directly after the [CLS] token. 
After the student response, we add either the question text or the question ID as input to the BERT model. Question text can help the model understand the question context and generalize across questions by leveraging their semantic relations. 
Question IDs enable the model to identify which question the target response belongs to, which can be helpful when the question text is not semantically meaningful; see Section~\ref{response} for an example. 
We can also add textual descriptions of the grading scales to the input. Since we use language models that are better at understanding text than numbers, we use ``bad, poor, fair, good, excellent'' to represent scores of 0, 1, 2, 3, and 4, respectively. 

Another key innovation is that, following recent approaches \cite{chen2021meta,min2021metaicl} for meta-training based in-context learning, we also input examples of scored responses, i.e., responses and corresponding scores, $(x^i_{\neg j}, y^i_{\neg j})$ from training dataset that belong to same question of the target response $x^i_j$. These examples provide further context to the model that the model can use to relate the target response to. Intuitively, when these examples are presented in the input, the AS model only needs to find example responses that are similar to the current response and use their scores to help score the current response. This task is easier for the AS model to learn than the real AS task when examples are not in the input. 

\section{Experiments}
This section details the experiments we conducted to validate our in-context meta-training approach for ASAG. Section \ref{dataset} discusses details on the real-world dataset student response dataset we use and how our pre-processing steps. Section \ref{metric} details evaluation metrics and baselines. We design three groups of experiments to test our approach's performance. In section \ref{response}, we examine how the approach performs on generalizing new student responses while having an assumption that the questions have already been seen during the training process. In section \ref{meta}, we examine the performance of our approach generalizing to scoring student response to new questions; in section \ref{ablation}, we run experiments to test which part of the in-context has the most significant impact on the performance of our approach.

\subsection{Dataset}
\label{dataset}

In this study, we use data collected from an online learning platform that has been used in prior work \cite{sami,erickson}. The dataset contains student responses to open-ended questions paired with scores provided by human graders. 
The dataset used in \cite{sami,erickson} consists of 141,612 total student responses from 25,069 students to 2,042 questions, scored by 891 different graders. The numeric score given to each response is in a 5-point scale from 0-4 with 4 as full credit and 0 as no credit. We refer to this dataset as $D_{orig}$. 

$D_{orig}$ contains some noisy data points that increase the difficulty of learning. First, some student responses are the same, but the teacher grades are different. Second, all corresponding student responses get full credit for some questions. For example, even the student's response is ``I do not know", the response's grade is still full credit. Third, some students' responses are answered by image, making the text content empty. Fourth, similar issues on question body; some questions do not have semantic meaning (such as questions that refer to a question in a book that we cannot access) or are represented as tables or images. For this work, we mainly focus on questions with corresponding students' responses and scoring the responses no matter which student is and who is grading. Thus we hope to reduce the effect of these noisy data points and further clean up the dataset. We found that some student responses are the same but the teacher grades are different; therefore, we re-label 2,130 inconsistent responses with the highest grade that the corresponding response text can get. We remove 8,835 student responses that contain only images or broken characters (non-English words, non-math terms). Since our in-context meta-learning approach needs to learn coherent information between questions using question text, we need high qualify question text. We remove responses (9,930 number of responses) with a question body (231 number of questions) that does not have semantic meaning. We also remove questions (478) that contain less than 25 number of students' responses. We called the new dataset $D_{clean}$, it contains 131,046 responses in total and 1,333 questions. Table \ref{tab:example} shows some examples data points of this dataset. For each data point, it contains the student response, problem text, problem id and teacher grade.  

\begin{table*}[h]
\caption{Example questions, student responses, and scores in the dataset.}
\label{tab:example}
\resizebox{1.8\columnwidth}{!}{
\begin{tabularx}{\textwidth}{ l|X|X|l }
  \toprule
  $q_{id}$ : question unique id & $q_{text}$ : question text & $x$ : student response & $y$ : teacher grade \\ \hline
   112348 & Write a function rule and a recursive rule for a line that contains the points (-4, 11), (5, -7), and (7, -11) &  Don't know what a recursive rule is & 0 \\ \hline

  32147 & Ryan had \$800 of his summer job earnings remaining when school started. He plans to use this amount as spending money throughout the 10 months of his school year. please indicate the 3 most important words/phrases in the question &  The 3 most important words or phrases in the question are \$800, 10, and months. & 4  \\ \hline
  
  32149 & Ryan will divide the \$800 into 10 equal amounts of \$80. If he completely spends \$80 during each month of his school year, how much of his earnings will remain at the end of the third month of his school year? Explain how you got your answer. & he will have \$560 left. 800-240=560 & 4 \\ 
  \bottomrule
\end{tabularx} }
\end{table*}

\subsection{Metrics}
\label{metric}

For the evaluation of math ASAG methods, we utilize three evaluation metrics for categorical, integer-valued scores, following prior work \cite{sami,erickson}. The first metric is area under the receiver operating characteristic curve (\textbf{AUC}), which is designed for binary classification problems. Instead, we calculate the AUC in a way similar to \cite{hand2001simple} by averaging the AUC numbers over each possible score category, treating them as separate binary classification problems. The second metric is the root mean squared error (\textbf{RMSE}) which simply treats the score categories as numerical values. The third and most important metric is the multi-class Cohen's \textbf{Kappa} that is often used for ordered categories, which fits the setting of our ASAG data. 

\subsection{Scoring new responses}
\label{response}
\subsubsection{Experimental Setting}
For this experiment, we focus on comparing the performance of our approach to baselines on generalizing to new responses. 
We randomly divide all example \textbf{responses} in $D_{orig}$ (we use this dataset for a fair comparison to \cite{sami,erickson}) into 10 equally-sized folds for cross validation. For each run, we use 8 folds for training, 1 fold for validation to select a training epoch with the best performance on this fold and 1 fold for the final testing of all methods. 
Under this setting, we ensure every question is contained in the training set so for every response in the test set, our models have seen scored response examples from the exact same question in the training set.

For our approach we use MathBERT \cite{mathbertcrap} as the pre-trained model with 110M parameters as the base scoring model \footnote{\url{https://huggingface.co/tbs17/MathBERT}}. We use the Adam optimizer, a batch size of 16, a learning rate of 1e-5 for 5 epochs on an NVIDIA RTX 8000 GPU. We do not perform any hyper-parameter tuning and simply use the default setting. For each training response, we randomly sample one in-context example per score class and fill up with the rest of training examples up to 25 in total from the training dataset for the corresponding question. Due to the restriction on input length for language models (512 for MathBERT), we truncate an example to a maximum of 70 tokens if necessary to ensure that the question, the target response to score, and all examples all fit in. For testing, we repeat the process of randomly sampling examples eight times for each target student response to be scored and average the predicted score class probabilities.  

We use an evaluation setting that follows from the one used in \cite{erickson}, for a fair comparison to compare it with SBERT-Canberra (SBERT-C) \cite{sami}, the current state-of-the-art method. The evaluation utilizes a 2-parameter Rasch model \cite{wright1977solving}; We include three groups of terms as covariates in the Rasch model: i) the student ability and question difficulty parameters, ii) the score category predictive probabilities according to the trained scoring method, and iii) the number of words in the response. After training the scoring model, we use the predicted scoring probabilities to learn regression coefficients and the ability/difficulty parameters. Intuitively, this evaluation setup studies how textual information in open-ended responses help \emph{in addition to} student ability and question difficulty during scoring; its purpose is not to evaluate how accurately response scoring models are themselves. 

For this evaluation, we use Problem ID as input for each training response to help the model adapt to the task. We do not use question text as input since $D_{orig}$ contains many (709 out of 2,042) question texts that have no semantic meaning (e.g., ``For Page 100 question b, answer the question''). This noisy question text cannot help the model recognize different questions and may confuse the model.

\subsubsection{Results and Analysis}

Table~\ref{tab:meta} shows the average value for all metrics across the 10 folds for our method (Meta In-context), the SBERT-C baseline, and other baselines studied in \cite{sami}. We see that our method is able to achieve a 0.02 (or 4.2\%) improvement over the best performing baseline, SBERT-C, on the most important metric, Kappa, while also outperforming on the other two metrics with smaller margins. This improvement validates the effectiveness of our overall method and further pushes the boundary on math ASAG. This improvement is more significant on the cleaned dataset $D_{clean}$, which we use for further evaluation next. We further note that there is a discrepancy between metric values (high AUC, low Kappa) on this experiment compared to other experiments due to the Rasch model-based setup.

\begin{table}[ht]
\centering
\caption{Evaluation results using the same dataset and under the same evaluation setting as \cite{erickson,sami} show that our scoring method outperforms existing methods.}
\scalebox{.9}{
\begin{tabular}{l|c|c|c}
\toprule
Model & AUC & RMSE & Kappa  \\
\midrule
Rasch* + Meta In-context (ours) & \textbf{0.861} & \textbf{0.541} & \textbf{0.496} \\
Rasch* + SBERT-Canberra & 0.856 & 0.577 & 0.476 \\
Baseline Rasch & 0.827 & 0.709 & 0.370 \\
Rasch + Number of Words & 0.825 & 0.696 & 0.382 \\
Rasch* + Random Forest & 0.850 & 0.615 & 0.430 \\
Rasch* + XGBoost & 0.832 & 0.679 & 0.390 \\
Rasch* + LSTM & 0.841 & 0.637 & 0.415 \\
\bottomrule
\end{tabular}}
\label{tab:meta}
\end{table}

\begin{table*}[ht]
\centering
\renewcommand{\arraystretch}{1.2} 
\caption{Ablation results for different design components of our method on $D_{clean}$. Most components contribute significantly.} \label{tab:abl}
\resizebox{1.6\columnwidth}{!}
{
\begin{tabular}{|c|c|c|c|c|c|c|c|c|}
\hline
\multicolumn{5}{|c|}{Method Component} &
\multicolumn{3}{c|}{Metric}\\ \hline
Question Text & Question ID & Scale & Example & MathBERT &
AUC &
RMSE &Kappa \\ \hline 
$\checkmark$ &
\multicolumn{1}{c|}{}& 
\multicolumn{1}{c|}{$\checkmark$}& $\checkmark$& $\checkmark$ & 
\multicolumn{1}{c|}{\textbf{0.733} $\pm 0.006$}&\multicolumn{1}{c|}{1.077 $\pm 0.002$} & 0.589 $\pm 0.004$\\ \hline
\multicolumn{1}{|c|}{}& 
\multicolumn{1}{c|}{$\checkmark$}& 
\multicolumn{1}{c|}{$\checkmark$}& $\checkmark$  & $\checkmark$& 
\multicolumn{1}{c|}{0.724 $\pm 0.007$}&\multicolumn{1}{c|}{1.083 $\pm 0.003$} & 0.585 $\pm 0.006$\\ \hline
\multicolumn{1}{|c|}{$\checkmark$}& 
\multicolumn{1}{c|}{}&
\multicolumn{1}{c|}{$\checkmark$}&& $\checkmark$& 
\multicolumn{1}{c|}{0.710 $\pm 0.006$}& \multicolumn{1}{c|}{1.278 $\pm 0.002$} & 0.568 $\pm 0.004$\\ \hline
\multicolumn{1}{|c|}{}& 
\multicolumn{1}{c|}{}& 
\multicolumn{1}{c|}{$\checkmark$}& $\checkmark$& $\checkmark$& 
\multicolumn{1}{c|}{0.720 $\pm 0.008$}&\multicolumn{1}{c|}{1.088 $\pm 0.001$} & 0.583 $\pm 0.009$\\ \hline
\multicolumn{1}{|c|}{$\checkmark$}& 
\multicolumn{1}{c|}{}& 
\multicolumn{1}{c|}{}& $\checkmark$& $\checkmark$&
\multicolumn{1}{c|}{0.719 $\pm 0.008$} & \multicolumn{1}{c|}{1.091 $\pm 0.003$} & 0.582 $\pm 0.005$\\ \hline
\multicolumn{1}{|c|}{$\checkmark$}&\multicolumn{1}{c|}{}& \multicolumn{1}{c|}{$\checkmark$}&$\checkmark$&&      \multicolumn{1}{c|}{0.731 $\pm 0.007$}&\multicolumn{1}{c|}{\textbf{1.051 $\pm 0.004$}} & \textbf{0.604 $\pm 0.010$} \\ \hline
\end{tabular} }
\end{table*}

\subsubsection{Ablation Study} 
\label{ablation}
We conduct an ablation study to verify the effectiveness of each component of our scoring method: using question text as input vs.\ using only question ID as input, adding textual instructions to provide information on the scoring scale, using scored examples to provide additional context, and using MathBERT as the base language model to fine-tune vs.\ using BERT. For this evaluation, we use the cleaned dataset $D_{clean}$ and a different experimental setting to directly evaluate the scoring accuracy of ASAG methods without using the Rasch model. The rest of the experimental settings, from cross-validation to model training, remain the same. Table~\ref{tab:abl} shows the results for all variants of our approach on all three metrics. We see that removing question text, textual instructions on scoring scale, and scored examples as input all result in significant degradation in scoring accuracy on some (or all) metrics. Specifically, removing scored examples results in the most significant drop in scoring accuracy, by around 0.02 in Kappa; this result validates the effectiveness that providing in-context examples can significantly benefit language models by helping them adapt to the current task (question). 
This result clearly validates our intuition that in-context examples reduce the difficulty of the AS task by changing the nature of the task from scoring to finding similar responses, which is easier. 
Removing question text also results in a (less significant) accuracy drop off: this result directly contradicts our observations in the previous experiment using the original dataset in \cite{sami,erickson} where we found that inputting the question text results in worse performance than inputting only the question ID. The likely reason for this result is that the cleaned dataset $D_{clean}$ contains much more questions that are semantically meaningful, which are helpful to include in the scoring method to provide important information on the scoring task.

A surprising but important result of this experiment is that using MathBERT results in a small drop off in performance (0.015 on Kappa, 0.007 on RMSE, and a 0.002 improvement on AUC) compared to using BERT. This observation is counter-intuitive since MathBERT is specifically designed to handle math expressions and trained on mathematical content, while BERT is not. To further examine why MathBERT underwhelms on the scoring task, we further investigate its performance on subsets of responses divided according to how much math information is contained in them. Specifically, we divided responses in the test set into two groups according to the amount of mathematical expressions involved: $D_{math}$ that contains responses where more than half of the tokens in the response are mathematical tokens and $D_{text}$ that contains the rest of the responses. Table~\ref{tab:groups} shows scoring accuracy for our approach using MathBERT and BERT as the base language model on these different response subsets. We see that on responses that are primarily textual, BERT outperforms MathBERT, which suggests that MathBERT loses some ability to encode textual information. On responses that are primarily mathematical, MathBERT performs similarly to BERT on RMSE and Kappa while outperforming BERT on AUC. This result suggests that MathBERT may have some benefit in handling mathematical tokens but the advantage may be minimal. Therefore, an important avenue for future work is to develop language models that are capable of representing and understanding mathematical content.

\begin{figure*}
\includegraphics[width=0.32\textwidth]{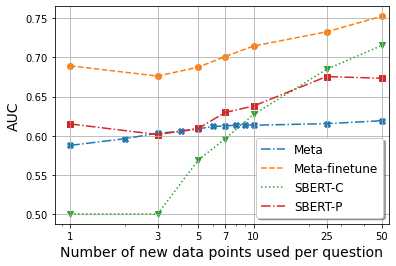}
\includegraphics[width=0.32\textwidth]{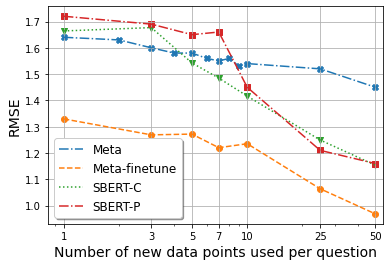}
\includegraphics[width=0.32\textwidth]{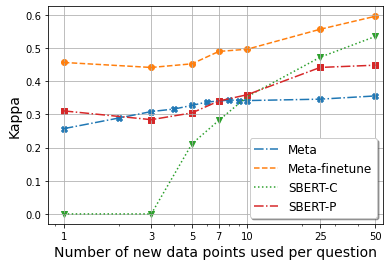}
        \caption{Results on generalizing to previously unseen questions using a few scored examples on all three metrics. Our approach, Meta-finetune, consistently outperforms SBERT-C and SBERT-P. Even without adjusting the model and using the scored examples as input (Meta), we outperform SBERT-C when the number of examples is small.}
        \label{fig:three graphs}
\end{figure*}

\begin{table}[ht!]
\caption{Scoring accuracy on responses that contain more mathematical tokens vs.\ more text tokens.} \label{tab:groups}
\resizebox{1\columnwidth}{!}
{
\begin{tabular}{cc|c|c|c}
\toprule
Data& approach&
\multicolumn{1}{c|}{AUC} & \multicolumn{1}{c|}{RMSE}&\multicolumn{1}{c}{Kappa}\\ 
\midrule
\multirow{2}{*}{D\_math} 
& MathBERT & \textbf{0.755} $\pm 0.008$& \textbf{0.587} $\pm 0.003$& 0.690$\pm 0.008$\\ 
& BERT& 0.741 $\pm 0.010$& 0.610 $\pm 0.009$& \textbf{0.691} $\pm 0.020$\\ \hline \\ \hline
\multirow{2}{*}{D\_text} 
& MathBERT & 0.713 $\pm 0.006$ & 1.022 $\pm 0.004$& 0.523 $\pm 0.006$\\ 
& BERT& \textbf{0.716} $\pm 0.006$ & \textbf{1.001} $\pm 0.003$& \textbf{0.542} $\pm 0.008$\\ 
\bottomrule
\end{tabular}
}
\end{table}

\subsection{Scoring new questions}
\label{meta}
\subsubsection{Experimental Setting}
For this experiment, we focus on testing the performance of our approach on generalizing to new questions (tasks) without seeing scored examples and how quickly our approach can adapt to them using few examples. Therefore, we randomly divide all questions in $D_{clean}$ into 5 equally-sized folds in terms of the number of questions instead of the number of responses. As a result, the number of responses in each fold may vary ($26,229 \pm 689$) since the number of student responses to each question is different. For each run, we use 4 folds for training and 1 fold for testing. 

On the test set, we make $n \in \{0, 1, 3, 5, 7, 10, 25, 50, 80\}$ scored responses per question available to methods trained on the training dataset and evaluate their ability to score other responses. 
We emphasize that there is no overlap between training responses and test responses for these previously unseen questions. 
We use two settings for our method. For the first setting, \textbf{Meta}, we do not further adjust the trained scoring model; instead, we only feed these responses and their scores, i.e., in-context examples, to the trained scoring model. For cases where $n<25$, we only feed in $n$ examples even though the method was trained with 25 examples. For cases where $n>25$, we follow randomly sample $25$ examples from the $n$ total examples as input, following the same setting above. 
This experimental setting can be seen as ``\emph{zero-shot}'' learning where we directly test how a scoring method trained on other questions works on new questions without observing any scored responses.  

For the second setting, \textbf{Meta-finetune}, we further fine-tune our trained method on the $n$ new scored responses per question. During this process, for each response as the scoring target, we use the other $n-1$ responses as in-context examples. 
This experimental setting can be seen as ``\emph{few-shot}'' learning where we test how quickly a scoring method trained on other questions can adapt to new questions. 

Since \textbf{SBERT-C} is the current state-of-the-art math ASAG method on this dataset, we use it as our baseline. According to \cite{sami}, it calculates similarities between the target response and other responses to the same question. Then it picks the score of the response with the highest similarity to the target response as its prediction, which means that it is not capable of zero-shot generalization to new questions. Therefore, we use $n$ scored examples on these new questions to train the scoring method and evaluate on the other responses. 
We emphasize again that in both the zero-shot and few-shot settings, the scored examples are excluded from performance valuations. 

We also use an additional baseline \cite{pardos}, which we refer to as \textbf{SBERT-P}. This method uses SBERT to encode responses and questions and feed the resulting representations to a classifier for predictions. This method also trains a single unified model across and is thus capable of zero-shot generalization to previously unseen questions. We use $n$ scored examples on these new questions for SBERT-P to train on to evaluate it in the few-shot learning setting.

\subsubsection{Result and Analysis}

\begin{table}[t!]
    \centering
    \caption{Scoring accuracy for different methods on generalization to new questions not seen during training, using a small number of scored examples.}
    \resizebox{\columnwidth}{!}{
        \begin{tabular}{p{1.5cm}|c|c|c|c}
            \toprule
            num-of new-data points / question & Method & \texttt{AUC} & \texttt{RMSE} & \texttt{KAPPA} \\
            \hline
            \multirow{3}{*}{0} &
            Meta & $0.533 \pm 0.017$ & $1.650 \pm 0.020$ & $0.100 \pm 0.052$ \\ 
            & SBERT-P & $0.558 \pm 0.006$ & $1.931 \pm 0.001$ & $0.170 \pm 0.013$ \\ 
            & SBERT-C & $-$ & $-$ & $-$ \\ \hline
            
            \multirow{4}{*}{1 } &
            Meta & $0.588 \pm 0.012$ & $1.641 \pm 0.013$ & $0.257 \pm 0.041$ \\ 
            & Meta-finetune & $0.689 \pm 0.033$ & $1.329 \pm 0.009$ & $0.456 \pm 0.048$ \\ 
            & SBERT-P & $0.615 \pm 0.022$ & $1.721 \pm 0.011$ & $0.310 \pm 0.043$ \\ 
            & SBERT-C &  $0.500 \pm 0.001$ & $1.664 \pm 0.009$ & $0.000 \pm 0.001$ \\ \hline
            
            \multirow{4}{*}{3} &
            Meta & $0.606 \pm 0.012$ & $1.620 \pm 0.013$ & $0.308 \pm 0.041$ \\ 
            & Meta-finetune & $0.676 \pm 0.010$ & $1.269 \pm 0.010$ & $0.441 \pm 0.017$ \\ 
            & SBERT-P & $0.601 \pm 0.040$ & $1.691 \pm 0.010$ & $0.284 \pm 0.071$ \\ 
            & SBERT-C &  $0.501 \pm 0.001$ & $1.677 \pm 0.009$ & $0.000 \pm 0.001$ \\ \hline
            
            \multirow{4}{*}{5 } &
            Meta & $0.589 \pm 0.013$ & $1.581 \pm 0.013$ & $0.289 \pm 0.043$ \\ 
            & Meta-finetune & $0.688 \pm 0.009$ & $1.272 \pm 0.013$ & $0.452 \pm 0.021$ \\ 
            & SBERT-P & $0.610 \pm 0.028$ & $1.650 \pm 0.010$ & $0.284 \pm 0.050$ \\ 
            & SBERT-C &  $0.569 \pm 0.061$ & $1.543 \pm 0.080$ & $0.211 \pm 0.016$ \\ \hline
            
            \multirow{4}{*}{7} &
            Meta & $0.611 \pm 0.011$ & $1.548 \pm 0.011$ & $0.341 \pm 0.040$ \\ 
            & Meta-finetune & $0.701 \pm 0.010$ & $1.220 \pm 0.008$ & $0.489 \pm 0.022$ \\ 
            & SBERT-P & $0.630 \pm 0.037$ & $1.662 \pm 0.012$ & $0.340 \pm 0.064$ \\ 
            & SBERT-C &  $0.569 \pm 0.006$ & $1.485 \pm 0.011$ & $0.282 \pm 0.019$ \\ \hline
            
            \multirow{4}{*}{10} &
            Meta & $0.614 \pm 0.010$ & $1.543 \pm 0.013$ & $0.342 \pm 0.043$ \\ 
            & Meta-finetune & $0.716 \pm 0.008$ & $1.235 \pm 0.009$ & $0.496 \pm 0.021$ \\ 
            & SBERT-P & $0.638 \pm 0.031$ & $1.453 \pm 0.018$ & $0.359 \pm 0.080$ \\ 
            & SBERT-C & $0.627 \pm 0.008$ & $1.416 \pm 0.009$ & $0.353 \pm 0.019$ \\ \hline
            
            
            
            \multirow{4}{*}{80} &
            Meta & $0.626 \pm 0.024$ & $1.550 \pm 0.016$ & $0.373 \pm 0.074$ \\ 
            & Meta-finetune & $0.765 \pm 0.010$ & $0.940 \pm 0.015$ & $0.636 \pm 0.042$ \\ 
            & SBERT-P & $0.704 \pm 0.033$ & $1.101 \pm 0.011$ & $0.523 \pm 0.020$ \\ 
            & SBERT-C & $0.735 \pm 0.017$ & $1.094 \pm 0.008$ & $0.581 \pm 0.042$ \\
            \bottomrule
        \end{tabular}
    }
    \label{table:level}
\end{table}
Table~\ref{table:level} shows the experimental results averaged over all folds. We see that Meta-finetune outperforms the other three approaches on all values of $n$ for all metrics, achieving satisfactory results of AUC = $0.689$, RMSE = $1.329$ and Kappa = $0.456$ in the one-shot learning setting ($n=1$), significantly outperforming Meta, SBERT-P and SBERT-C (by up to 50\% on Kappa).  
The performance of Meta-finetune stabilizes as $n$ increases and still outperforms SBERT-C ($0.03$ on AUC, $0.154$ on RMSE and $0.055$ on Kappa) and SBERT-P ($0.11$ on AUC, $0.161$ on RMSE and $0.113$ on Kappa) at $n=80$. These results clearly demonstrate that, compared to SBERT-C and SBERT-P, our method is highly effective at ``warm-starting'' scoring models on new questions since it is able to get a sense of how responses should be scored from scored responses to other questions. Again, we note that in-context examples changes the nature of the task from AS to finding similar responses; as a result, models can learn this task quicker and adapt to new questions using only a few examples.

SBERT-C, on the other hand, can barely work in few-shot learning settings, i.e., $n \in \{1, 3\}$. This observation is not surprising since SBERT-C does learn a scoring model from scratch and cannot work when the number of training data points is less than the number of possible score categories. The performance of SBERT-C starts to gradually increase when $n>5$ but is still significantly worse than Meta-finetune. 

Meta, the method for zero-shot learning, although fails to generalize well (only $0.533$ in AUC and $0.1$ in Kappa at $n=0$) without seeing any training data, still significantly outperforms SBERT-C with $n \in \{1, 3\}$ and performs similarly to SBERT-P. This advantage only disappears at $n=10$. To further illustrate this difference, we plot the three metrics vs.\ $n$ for all methods in Figure~\ref{fig:three graphs}. We see that Meta's AUC and Kappa values are higher than that for SBERT-C until $n$ reaches around 8, which indicates that even without re-training the model, it is more suitable for few-shot learning than SBERT-C on new questions. 

\subsection{Qualitative Error Analysis}
In this section, we qualitatively analyze the prediction errors made by our ASAG method. We use the model trained on $D_{clean}$, with problem text + scale + examples as input into MathBERT for our analysis.

\subsubsection{Feature analysis}
To analyze the difference between correct predictions and incorrect predictions, we extract several features that capture properties of the questions and responses to better understand the strengths and weaknesses of the trained ASAG method. As shown in Table~\ref{tab:feature}, ``Response math tokens'' represents the percentage of math tokens in a response; ``Response contains img/table'' represents whether a response has images or tables; ``Response length'' represents the number of tokens a response; ``score'' represents the actual score given by the graders;  ``Number of graders'' represents the number of graders that graded each response to the question; ``question length'' represents the number of tokens the corresponding question has and ``question math tokens'' represents the percentage of math tokens in the question.

\begin{table}[ht]
\centering
\caption{Features analysis between correct predictions and incorrect predictions. * means the difference is significant ($p\_value < 0.005$).}
\scalebox{.78}{
\begin{tabular}{lcc}
\toprule
Features (avg.) & Correct Prediction & Incorrect Prediction  \\
\midrule
Response math tokens (\%)*  &30.6&25.1\\ 
Response contain &&\\ img/table (\%)* &1.29&2.88\\ 
Response length*  &17.4&29.5\\ 
Score* &3.25&2.13 \\ 
Number of graders &2.53&2.48 \\ 
Question length* &37.1&39.1 \\ 
Question math tokens (\%) &8.12&7.31\\ 
\bottomrule
\end{tabular}}
\label{tab:feature}
\end{table}

We observe a significant difference ($p\_value < 0.005$) between values of the correct predictions and values of the incorrect predictions. We make the following observations:
\begin{itemize}
	\item The scoring method is more accurate at scoring responses with higher percentage of math tokens and it becomes less accurate when there are higher percentage of plain texts. 
	\item The scoring method is more accurate when the response contains images or tables that words can not represent. 
	\item The scoring method is more accurate at scoring shorter responses.
	\item The scoring method is more accurate at scoring responses with shorter question description.
 	\item The scoring method is more accurate when the average score of the response is higher. This observation indicates that the model is better at scoring responses with higher quality.
 	\item There is no obvious distinction in grading accuracy for responses with different numbers of graders or to questions with different math tokens percentages. 
\end{itemize} 

\subsubsection{Question topic and type error analysis}
Table 8 lists the summarization of scoring accuracy on different question topics and types. We extract the topics and types from question text using BERTopic \cite{bertopic}. BERTopic is a topic modeling technique that leverages transformers and term and document frequencies \cite{data-mining} to create easily interpretable topics. Overall, we see that the trained scoring method has better Kappa scores on questions that are primary text-based or involve equations. The result is not surprising since we adapted MathBERT, which likely sees many text-based questions during its pre-training stage. Questions that require students draw graphs in their response also have high Kappa scores; however this result is mainly due to the fact that most of these responses are given full credit, making them easy for scoring methods to make predictions. 
On the other hand, the trained scoring method has worse Kappa scores on estimation-type and (only a few) multiple-choices questions. This observation can be explained by language models not being trained to capture number sense and thus struggle at numerical reasoning \cite{numgpt}. For multiple-choice questions, the response, i.e., the multiple-choice option, is semantically meaningless, which does not provide meaningful context to the scoring method.

\begin{table}[ht]
\caption{Scoring accuracy on different question topics and types. Results are shown in increasing order of the Kappa score. * means the score is better than the average across all responses.}
\resizebox{\columnwidth}{!}{
\begin{tabular}{cclll}
\toprule
Topic& Type &   AUC &  RMSE &  Kappa    \\
\midrule
Misc. & Multiple-choice        & 0.631   & 1.472  &  0.400  \\
Math& Table calculation     & 0.659   & 1.345  &  0.445  \\
Algebra& Estimation              & 0.702   & 1.310  &  0.536  \\
Calculus& Estimation             & 0.716   & 1.241  &  0.546  \\
Algebra& Table creation         & 0.731   & 0.823* &  0.606*  \\
Algebra& Equation writing        & 0.732   & 1.023  &  0.612* \\
Algebra& Graph drawing            & 0.734*  & 0.725* &  0.629*  \\
Math& Word question             & 0.735*  & 0.663* &  0.647*  \\
Calculus& Graph drawing           & 0.736*  & 0.610* &  0.758*  \\
\bottomrule
\end{tabular}
}
\label{tab:question type}
\end{table}

\begin{table*}[]
\caption{Examples of scoring errors made by our trained method.}
\centering
\resizebox{1.4\columnwidth}{!}{
\begin{tabular}{|cccc|}
\hline
\multicolumn{4}{|l|}{\begin{tabular}[c]{@{}l@{}} \textbf{Question:} Chelsea collects butterfly stickers. The picture shows how she placed them. \\Write a division sentence to show how she equally grouped her stickers. \_\_÷ \_\_ = \_\_\end{tabular}}      \\ 
\multicolumn{4}{|l|}{ \textbf{Most frequency correct response :} 15/3=5} \\ \hline
\multicolumn{1}{|c|}{Error type} & \multicolumn{1}{c|}{Response}& \multicolumn{1}{c|}{Grade} & Predict \\ \hline
\multicolumn{1}{|c|}{\multirow{3}{*}{\begin{tabular}[c]{@{}c@{}}Poor reasoning \\ on math operator \\ and numerical token\end{tabular}}} 
& \multicolumn{1}{c|}{15* 5=3}& \multicolumn{1}{c|}{2} & 4 \\ \cline{2-4} 
\multicolumn{1}{|c|}{}& \multicolumn{1}{c|}{5/3=15}  & \multicolumn{1}{c|}{0} & 2  \\ \cline{2-4} 
\multicolumn{1}{|c|}{}& \multicolumn{1}{c|}{15\_3=12} & \multicolumn{1}{c|}{1}     & 2       \\ \hline
\multicolumn{1}{|c|}{Spelling error}  & \multicolumn{1}{c|}{3 times 5 eques 15} & \multicolumn{1}{c|}{4}  & 2  \\ \hline
\multicolumn{1}{|c|}{\multirow{2}{*}{\begin{tabular}[c]{@{}c@{}}Confused by \\ paraphrased responses \end{tabular}}}& \multicolumn{1}{c|}{I think that it is 5x3=15} & \multicolumn{1}{c|}{4}     & 1       \\ \cline{2-4} 
\multicolumn{1}{|c|}{} & \multicolumn{1}{c|}{\begin{tabular}[c]{@{}c@{}}she place them like in 3 groups \\ and she even did the answer \\ but she did not new the each group\end{tabular}} & \multicolumn{1}{c|}{2} & 0 \\ \hline
\multicolumn{1}{|c|}{Meaningless response} & \multicolumn{1}{c|}{see attachment} & \multicolumn{1}{c|}{3}& 0 \\ \hline 
\end{tabular}}
\label{tab:error-type}
\end{table*}

\subsubsection{Error type analysis}

For this analysis, we choose a question with scoring accuracy below the average on our dataset to analyze the types of errors made by our trained scoring method. Table~\ref{tab:error-type} shows selected responses with erroneous score predictions and the types of these errors. The question asks students to write an equation with a popular correct response $15 / 3 = 5$. We make the following observations on typical error types (apart from some obvious human grader errors, which we omit): 
\begin{itemize}
	\item The first error type indicates that our trained scoring method can still struggle on mathematical reasoning and handling numerical tokens. The incorrect responses ``15*5=3'' and ``5/3=15'' have the same numerical tokens but with different ordering and an incorrect operator token compared to the correct response, which completely changes their meaning.  The trained scoring method tends to overestimate their scores. This observation suggests that we need base language models with stronger numerical reasoning abilities. 

    \item The second error type indicates that our trained scoring method can struggle with spelling errors in student responses. When the word ``equals'' is spelled incorrectly in a student response, it does not affect the human grader's ability to understand the student's intention. However, the trained scoring method puts a penalty on this spelling error.  

    \item The third error type indicates that our trained scoring method may not recognize paraphrased responses. As shown in the examples, student may add text such as ``I think that it is'' which does not alter the meaning of the response; however, it adds noise and  misled the prediction. 
    
    \item The fourth error type indicates that our trained scoring method cannot handle responses in unparsable format such as an attachment. 
\end{itemize}

\section{Conclusions and Future Work}

In this paper, we have proposed a language model fine-tuning-based method for automatic short answer grading for open-ended, short-answer math questions. Our method has two main components: a base MathBERT model pre-trained with educational content on math subjects, and a meta-learning-based, in-context fine-tuning method that promotes generalization to new questions with a carefully designed input format. Experimental results on a large real-world student response dataset revealed surprisingly contradicting findings: Using MathBERT instead of regular BERT, which is not trained on mathematical content, results in a decrease in scoring accuracy, while the in-context fine-tuning method results in significantly improved scoring accuracy compared to existing methods, especially on previously unseen questions. 

There are plenty of avenues for future work. First, the observation that MathBERT \cite{mathbertcrap} cannot outperform BERT as the base language model suggests that there is a need to develop more effective models for mathematical language. One promising direction is perhaps taken by another simultaneously proposed version of MathBERT \cite{mathbertgood} that leverages the inherent tree structure of mathematical expressions. Moreover, the noisiness of human grading that we observed in our experiments suggests that there is a need to develop ASAG methods that take inter-rater agreement into account \cite{whitehill2009whose}. 

Second, there is a need to further improve the completeness of the context information we provide to the base language model. Several possible sources of additional contextual information include the grade level of the question, the common core standard codes, and mathematical skill/concept tags, which can all provide information on the level of the question. Additionally, we may even directly incorporate relevant mathematical content into the model's input, e.g., by retrieving content chunks in textbooks or online resources using information retrieval methods \cite{drqa}. However, a potential challenge that needs to be resolved is how to concisely pack all relevant contextual information into the model without exceeding the input length limit of language models (usually 512 tokens). 

Third, in order to make ASAG methods more applicable in realworld educational scenarios, there is a need to thoroughly study the fairness aspects of these methods and ensure all students are treated fairly. There is a need to investigate how ASAG methods performs on different student populations; recent work has raised the concern that it is not clear that whether one should explicitly incorporate student demographic information during model training \cite{yu2021should}. Future work should explore how to incorporate fairness regularization into the training objective to promote methods that are fair across students \cite{Agarwal:Reduction:2018,Russell:World:2017,Zafar:Fairness:2017,Zemel:Learning:2013, fairness}.

\section{Acknowledgement}
The authors would like to thank the National Science Foundation for their support through grant IIS-2118706, NSF (e.g., 2118725, 2118904, 1950683, 1917808, 1931523, 1940236, 1917713, 1903304, 1822830, 1759229, 1724889, 1636782, \& 1535428), IES (e.g., R305N210049, R305D210031, R305A170137, R305A170243, R305A180401, \& R305A120125), GAANN (e.g., P200A180088 \& P200A150306), EIR (U411B190024 \& S411B210024), ONR (N00014-18-1-2768), and Schmidt Futures. None of the opinions expressed here are that of the funders.  Neil is funded under an NHI grant (R44GM146483) with Teachly as a SBIR.

%
\bibliographystyle{abbrv}
\bibliography{references}  

\balancecolumns
\end{document}